%% file: root.tex
\documentclass[twocolumn,journal]{IEEEtran}
\usepackage[latin9]{inputenc}
\usepackage{color}
\usepackage{amsmath}
\usepackage{amssymb}
\usepackage{graphicx}
\usepackage[unicode=true,
 bookmarks=true,bookmarksnumbered=true,bookmarksopen=true,bookmarksopenlevel=1,
 breaklinks=false,pdfborder={0 0 0},pdfborderstyle={},backref=false,colorlinks=false]
 {hyperref}
\hypersetup{pdftitle={Your Title},
 pdfauthor={Your Name},
 pdfpagelayout=OneColumn, pdfnewwindow=true, pdfstartview=XYZ, plainpages=false}

\makeatletter

\providecommand{\tabularnewline}{\\}

\let\oldforeign@language\foreign@language
\DeclareRobustCommand{\foreign@language}[1]{%
  \lowercase{\oldforeign@language{#1}}}

\usepackage[caption=false,font=footnotesize]{subfig}
\usepackage{cite}

\makeatother

\begin{document}
\input{macros.tex}

\title{\textcolor{black}{SmartArm: Suturing Feasibility of a Surgical Robotic
System on a Neonatal Chest Model}}
\author{Murilo~M.~Marinho,~\IEEEmembership{Member,~IEEE,} Kanako~Harada,~\IEEEmembership{Member,~IEEE,}
Kyoichi~Deie, Tetsuya~Ishimaru, and~Mamoru~Mitsuishi,~\IEEEmembership{Member,~IEEE}\thanks{This research was funded in part by the ImPACT Program of the Council
for Science, Technology and Innovation (Cabinet Office, Government
of Japan) and in part by the JSPS KAKENHI Grant Number JP20K20489.}\thanks{Murilo M. Marinho, Kanako Harada, and Mamoru Mitsuishi are with the
Department of Mechanical Engineering, The University of Tokyo, Tokyo,
Japan, e-mail \protect\href{http://\%7Bmurilo,\%20kanako,\%20mamoru\%7D@nml.t.u-tokyo.ac.jp}{\{murilo, kanako, mamoru\}@nml.t.u-tokyo.ac.jp}.}\thanks{Kyoichi Deie is with the Department of Pediatric Surgery, Kitasato
University, Kanagawa, Japan, e-mail: \protect\href{http://k.deie@med.kitasato-u.ac.jp}{k.deie@med.kitasato-u.ac.jp}.}\thanks{Tetsuya Ishimaru is with the Saitama Children\textquoteright s Medical
Center, Saitama, Japan e-mail: \protect\href{http://tetsuyaishimaru@gmail.com}{tetsuyaishimaru@gmail.com}.}}
\markboth{ACCEPTED FOR PUBLICATION IN THE IEEE TRANSACTIONS ON MEDICAL ROBOTICS
AND BIONICS}{Murilo M. Marinho \MakeLowercase{\emph{et al.}}: SmartArm: Suturing
Feasibility on a Neonatal Chest Model using a Versatile Surgical Robotic
System}
\maketitle
\begin{abstract}
Commercially available surgical-robot technology currently addresses
many surgical scenarios for adult patients. This same technology cannot
be used to the benefit of neonate patients given the considerably
smaller workspace. Medically relevant procedures regarding neonate
patients include minimally invasive surgery to repair congenital esophagus
disorders, which entail the suturing of the fragile esophagus within
the narrow neonate cavity. In this work, we explore the use of the
SmartArm robotic system in a feasibility study using a neonate chest
and esophagus model. We show that a medically inexperienced operator
can perform two-throw knots inside the neonate chest model using the
robotic system.
\end{abstract}


\section{Introduction}

\IEEEPARstart{N}{eonate} patients cannot currently benefit from commercial
surgical-robot technology, which mostly targets adult patients \cite{Thakre2008,Cundy2015,Takazawa2018}.
Medically significant applications for neonatal patients that could
highly benefit from robotic assistance include the treatment of congenital
disorders in the esophagus, two of those being tracheoesophageal fistula
(TEF) and esophageal atresia (EA).

TEF is a disorder in which the newborn has a connection between the
esophagus and the trachea and those must be surgically separated.
EA is a condition in which the esophagus of the newborn is formed
in two disconnected parts that have to be surgically reconnected.
The most challenging step in both surgical procedures is to suture
or anastomose the tiny esophagus/trachea of the neonate patient. In
general, this part of the procedure is viewed through a 4-mm-diameter
endoscope, and the suture is handled by a 3-mm-diameter forceps and
a 3-mm-diameter needle driver \cite{Deie2016}.

The constrained workspace in neonatal patients is a technical challenge
that has to be addressed in many fronts. Advancements in anatomically
proper chest and esophagus models for pediatric surgical training
\cite{Harada2016} are an important element to support the development
of surgical robot technology and provide an ethical benchmark to the
evaluation of surgical skill.

Our group and collaborators have been using such models to support
the development of technologies that com\textcolor{black}{pose the
SmartArm surgical system \cite{Marinho2020}. Th}e technologies include
the development of thin dexterous robotic instruments \cite{Arata2019},
suitable teleoperation algorithms \cite{Marinho2019}, and collision
avoidance strategies \cite{Marinho2018}. The SmartArm has been validated
in an endonasal procedure using the head model of an adult patient
\cite{Marinho2020}, and in prior related work we showed a proof-of-concept
of the underlying technologies required for robotic assistance in
procedures targeting infants \cite{Marinho2019}.

\begin{figure}[t]
\begin{centering}
\includegraphics[width=1\columnwidth]{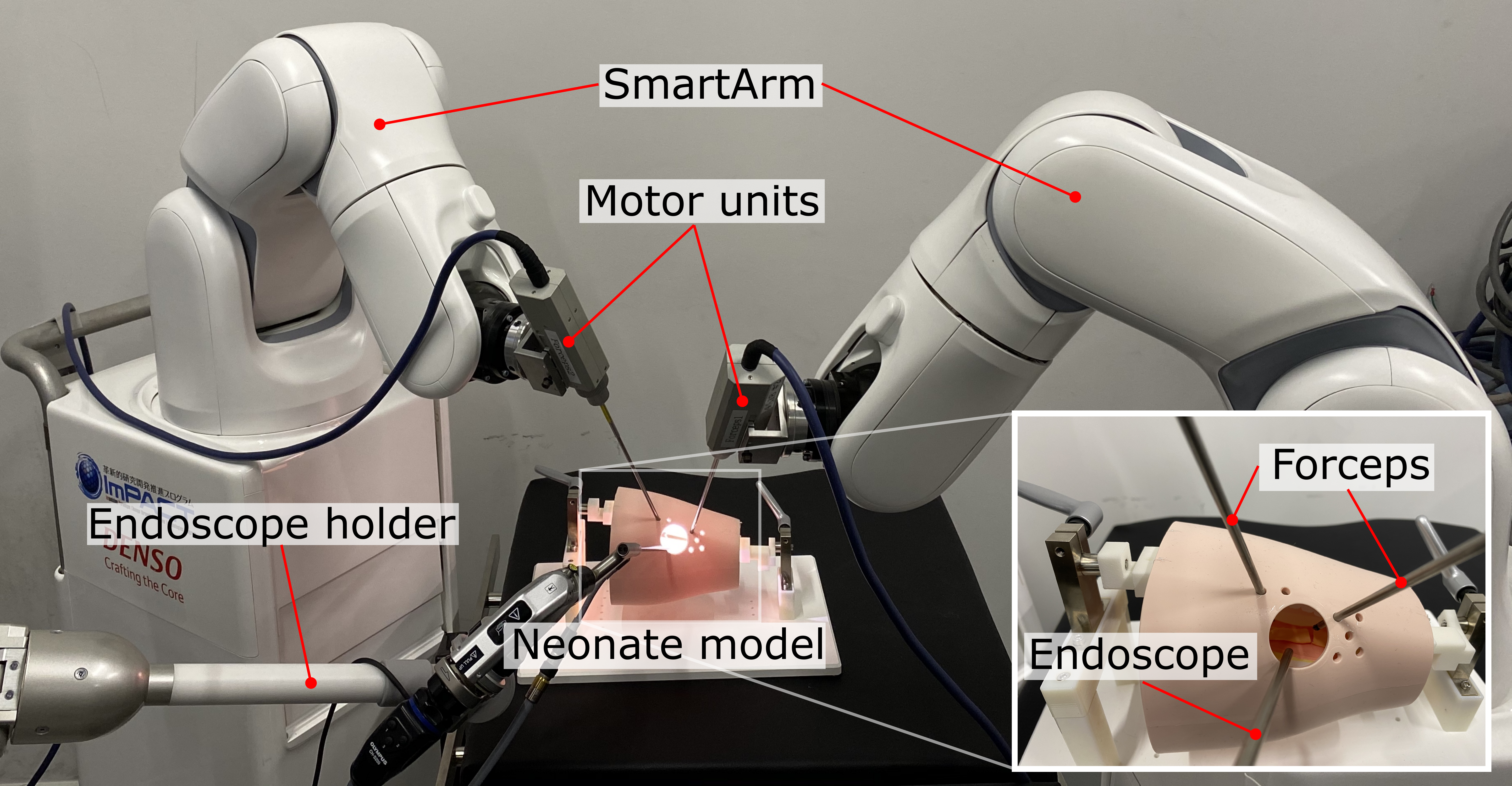}
\par\end{centering}
\caption{\label{fig:setup}The experimental setup for the esophageal suturing
using the SmartArm surgical system.}
\end{figure}

\paragraph*{Statement of contribution}

Based on those initial results, in this work, we investigate the feasibility
of using the SmartArm to suture an esophageal model developed for
the training of TEF and EA surgeries. To the best of the authors'
knowledge, intracorporeal suturing in the context of TEF and EA for
neonatal patients have not yet been addressed by robotic surgery.

\section{Materials and methods}

\subsection{Neonate surgical setup}

The robotic setup can be divided into the\emph{ patient side} (PS)
and the \emph{operator side} (OS), explained as follows.

\subsubsection{Patient side}

\textcolor{black}{Each six-degrees-of-freedom (6-DoF) robotic manipulator
(VS050, Denso Wave, Japan) that compose the SmartArm system were placed
at opposing sides of a neonate model. Each of those robotic manipulators
was equipped with a 3-DoF 20-cm-long 3.5-mm-diameter instrument \cite{Arata2019}.
Hence, each of the robots, $R_{1},R_{2}$, had nine DoF. In addition,
the view of the workspace was obtained through a $0^{\circ}$ rigid
endoscope}\footnote{\textcolor{black}{Rigid endoscopes can have a prism attached to their
distal end to incline the direction of view. The choice of angle is
based on the task and on the user.}}\textcolor{black}{{} held by an endoscope holder (Endoarm, Olympus,
Japan). An overview of the PS is shown in Fig.~\ref{fig:setup}.}

\subsubsection{\textcolor{black}{Operator side}}

\textcolor{black}{The OS of the setup was composed of two haptic interfaces
(Phantom Premium, 3DSystems, USA) with a customized stylus \cite{kamei2013master}.
Each of those interfaces sent the desired pose signals to one of the
robots in the PS.}

\subsection{\textcolor{black}{Control strategy}}

\textcolor{black}{The control strategy used in this work is a particular
case of the control strategy introduced in \cite[ Section IV]{Marinho2020},
and is divided into a quadratic-programming (QP)-based controller
on the PS and a Cartesian impedance controller on the OS. The control
strategy is explained in Sections~\ref{subsec:control_strategy_PS}
and \ref{subsec:Operator-sidesubsec:control_strategy_OS}.}

\subsubsection{\textcolor{black}{Patient side\label{subsec:control_strategy_PS}}}

Let the vector of joint configurations for each robot in the PS be
given by $\myvec q_{i}\in\mathbb{R}^{9}$ for $i=1,2.$ The PS control
is based on sending at each time step the joint-velocity reference
signal\footnote{In optimization literature, a minimizer is usually denoted with a
superscript asterisk, such as $\dot{\myvec q}_{i}^{*}$. Here we use
the superscript $o$ to avoid notational confusion with the quaternion
conjugate.} $\dot{\myvec q}_{i}^{o}$ to the joint-velocity controllers of each
robot. The joint velocities are obtained by solving\footnote{The solution of QP problems is obtained through a numerical solver.}
at each time step and in parallel for each robot the following QP
problem \textcolor{black}{
\begin{alignat}{1}
\dot{\myvec q}_{i}^{o}\in\underset{\dot{\myvec q}_{i}}{\text{arg min}}\  & \alpha f_{t,i}+\left(1-\alpha\right)f_{r,i}+f_{\Lambda,i}\label{eq:quadratic_problem_teleoperation}\\
\text{subject to}\  & \mymatrix W_{i}\dot{\myvec q}_{i}\preceq\myvec w_{i},\nonumber 
\end{alignat}
where $f_{t,i}\triangleq\norm{\mymatrix J_{i,t}\dot{\myvec q}_{i}+\eta\vecthree{\tilde{\quat t}_{i}}}_{2}^{2}$,
$f_{r,i}\triangleq\norm{\mymatrix J_{i,r}\dot{\myvec q}_{i}+\eta\vecfour{\tilde{\myvec r}_{i}}}_{2}^{2}$,
and $f_{\Lambda,i}\triangleq\norm{\mymatrix{\Lambda}\dot{\myvec q}_{i}}_{2}^{2}$
are the cost functions related to the end-effector translation}\footnote{\textcolor{black}{The translation is written as a pure quaternion
in the form $\quat t=t_{x}\imi+t_{y}\imj+t_{z}\imk$ with $t_{x},t_{y},t_{z}\in\mathbb{R}$
and $\imi^{2}=\imj^{2}=\imk^{2}=\imi\imj\imk=-1$. Moreover, $\vecthree{\quat t=\begin{bmatrix}t_{x} & t_{y} & t_{z}\end{bmatrix}^{T}\in\mathbb{R}^{3}}.$}}\textcolor{black}{, $\quat t_{i}$; the end-effector rotation}\footnote{\textcolor{black}{The rotation is written as a unit quaternion in
the form $\quat r=\cos\left(\phi/2\right)+\quat v\sin\left(\phi/2\right)$
where $\phi\in\mathbb{R}$ is the angle of rotation about the unit-axis
defined by the pure quaternion $\quat v=v_{x}\imi+v_{y}\imj+v_{z}\imk$.
In addition, $\vecfour{\quat r}=\begin{bmatrix}\cos\left(\phi/2\right) & \sin\left(\phi/2\right)v_{x} & \sin\left(\phi/2\right)v_{y} & \sin\left(\phi/2\right)v_{z}\end{bmatrix}^{T}\in\mathbb{R}^{4}.$
Note that $\quat r^{*}=\cos\left(\phi/2\right)-\quat v\sin\left(\phi/2\right)$
is the conjugate of $\quat r$ and represents the inverse rotation.}}\textcolor{black}{, $\quat r_{i}$; and joint velocities of the $i$-th
robot, respectively. In addition, each $i$-th robot has a translation
Jacobian $\mymatrix J_{i,\quat t}\in\mathbb{R}^{3\times9}$, a translation
error $\tilde{\quat t}_{i}\triangleq\quat t_{i}-\quat t_{i,d}$, a
rotation Jacobian $\mymatrix J_{i,\quat r}\in\mathbb{R}^{4\times9}$,
and a switching rotational error $\tilde{\quat r}_{i}\left(\quat r_{i},\quat r_{i,d}\right)\triangleq\tilde{\quat r}_{i}$
given by
\[
\tilde{\quat r}_{i}\triangleq\begin{cases}
\conj{\left(\myvec r_{i}\right)}\myvec r_{i,d}-1 & \text{if }\norm{\conj{\myvec r_{i}}\myvec r_{i,d}-1}_{2}<\norm{\conj{\myvec r_{i}}\myvec r_{i,d}+1}_{2}\\
\conj{\left(\myvec r_{i}\right)}\myvec r_{i,d}+1 & \text{otherwise},
\end{cases}
\]
based on the dual quaternion invariant error \cite{Figueredo2013},
where $\quat r_{i,d}$ and $\quat r_{i}$ are the desired and current
end-effector orientations, respectively. The switching error is used
to circumvent the unwinding problem given that the group of unit quaternions
double covers $\mathrm{SO}\left(3\right)$ \cite{Siciliano2009}.
This way, if $\conj{\left(\myvec r_{i}\right)}\myvec r_{i,d}$ is
closer to $1$, the error is given by $\conj{\left(\myvec r_{i}\right)}\myvec r_{i,d}-1$;
conversely, if $\conj{\left(\myvec r_{i}\right)}\myvec r_{i,d}$ is
closer to $-1$, the error is given by $\conj{\left(\myvec r_{i}\right)}\myvec r_{i,d}+1$.
The desired translation signals, $\quat t_{i,d}$, and rotational
signals, $\quat r_{i,d}$, are defined by the operator handling the
haptic interfaces in the OS. Furthermore, $\eta\in\left(0,\infty\right)\subset\mathbb{R}$
is a proportional gain to reduce the task error and $\mymatrix{\Lambda}\in\mathbb{R}^{9\times9}$
is a positive definite damping matrix in the form
\[
\mymatrix{\Lambda}\triangleq\begin{bmatrix}\lambda_{R}\mymatrix I_{6\times6} & \mymatrix 0_{6\times3}\\
\mymatrix 0_{3\times6} & \lambda_{F}\mymatrix I_{3\times3}
\end{bmatrix},
\]
where $\lambda_{R},\lambda_{F}\in\left[0,\infty\right)\subset\mathbb{R}$
are used to balance the costs of using joints in the robot and in
the forceps, and $\mymatrix I$ and $\mymatrix O$ stand for the identity
and zero matrices, respectively. Lastly, $\alpha$ $\in$ $\left[0,1\right]\subset\mathbb{R}$
is a weight used to define the (soft) priority between the translation
and the rotation. The parameter $\alpha$ only visibly affects the
system when the robot is unable to }\textcolor{black}{\emph{simultaneously}}\textcolor{black}{{}
achieve the desired error convergence rate for both rotation and translation
owing to task-space or joint-space constraints. The parameter is often
found empirically for a given user and a given task \cite{Marinho2019,Marinho2020}.
This gives the designer more flexibility than resorting to nested
optimization algorithms \cite{Escande2014}, which would also increase
the computational time.}

\textcolor{black}{The linear constraints are used in the following
way
\begin{equation}
\overbrace{\begin{bmatrix}\mymatrix W_{JL,i}\\
\mymatrix W_{ES,i}
\end{bmatrix}}^{\mymatrix W_{i}}\dot{\myvec q}_{i}\preceq\overbrace{\begin{bmatrix}\myvec w_{JL,i}\\
\myvec w_{ES,i}
\end{bmatrix}}^{\myvec w_{i}},\label{eq:constraints}
\end{equation}
where $\mymatrix W_{JL,i}\triangleq\begin{bmatrix}-\mymatrix I_{9\times9} & \mymatrix I_{9\times9}\end{bmatrix}^{T}$
and $\myvec w_{JL,i}\triangleq\begin{bmatrix}\left(\myvec q_{\text{min},i}-\quat q_{i}\right)^{T} & \left(\myvec q_{\text{max},i}-\quat q_{i}\right)^{T}\end{bmatrix}^{T}$
are used to maintain the joint values within their lower and upper
bounds, given by $\myvec q_{\text{min},i}\in\mathbb{R}^{9}$ and $\myvec q_{\text{max},i}\in\mathbb{R}^{9}$
respectively \cite{cheng1994}.}

\textcolor{black}{For procedures in the constrained workspace inside
neonate patients, medical doctors have to take advantage of the natural
compliance of the patient's skin. This constraint is called an entry-sphere
as represented in Fig.~\ref{fig:model_inside}. To generate an entry-sphere
constraint for each robotic instrument, we used $\mymatrix W_{ES,i}\triangleq\distancejacobian{l_{i}}{p_{i}}{}$
and $\myvec w_{ES,i}\triangleq\eta_{d}(D_{\text{safe}}-D_{\text{ES},i})$,
where $\distancejacobian{l_{i}}{p_{i}}{}\in\mathbb{R}^{1\times9}$
is the line-to-point distance Jacobian \cite[Eq. 34]{Marinho2018}
relating the distance between the center-line of the shaft of the
instrument of the $i-$th robot and the point representing the center
of the entry-sphere for that instrument. In addition, $D_{\text{ES},i}\left(\myvec q_{i}\right)\triangleq D_{\text{ES},i}$
is the square-distance between the center of the entry-sphere and
the centerline of the shaft of the instrument \cite[Eq. 33]{Marinho2018}
and $D_{\text{safe}}\in\mathbb{R}^{+}$ is the maximum allowed square
distance. Lastly, $\eta_{d}\in\left(0,\infty\right)\subset\mathbb{R}$
is a gain that limits the velocity in which the robot can move towards
the boundaries of the entry-sphere constraint.} This technique provides
smooth collision avoidance and guarantees that the constraints will
not be violated \cite{Marinho2018}.

\paragraph*{\textcolor{black}{Stability}}

\textcolor{black}{The Lyapunov stability of linearly-constrained optimization-based
controllers was studied initially in \cite{gonccalves2016parsimonious}
and proven in more detail for strictly convex objective functions
as well in \cite{Junior2018}.}

\paragraph*{Existence and uniqueness of a solution}

There will always be a solution to Problem~\ref{eq:quadratic_problem_teleoperation}
as long as the constraints are not initially violated because this
means that $\dot{\myvec q}_{i}^{o}=\myvec 0$ is always a feasible
solution. This premise is valid for the present surgical robotic application
given that the robot will be safely placed inside the patient without
violating the constraints during the initial setup. Related literature
indicates that the solution of Problem~\ref{eq:quadratic_problem_teleoperation}
will always be unique if $\mymatrix{\Lambda}\succ0$ \cite{Escande2014,Junior2018},
because the objective function becomes strictly convex.

\subsubsection{Operator side\label{subsec:Operator-sidesubsec:control_strategy_OS}}

In order to provide haptic guidance to the operator, we add the standard
Cartesian force feedback on the OS for each haptic interface. The
Cartesian force feedback is proportional to the current task-space
error of the robots in the PS, in the form
\begin{align}
\quat{\Gamma}_{i,\text{OS}} & \triangleq-\eta_{f}\tilde{\myvec t}_{i}^{\text{OS}}-\eta_{V}\dot{\myvec t}_{i,\text{OS}},\label{eq:cartesian_impedance}
\end{align}
where $\quat{\Gamma}_{i,\text{OS}}$ is the force feedback on the
OS, $\eta_{f},\eta_{V}$ $\in$ $\left(0,\infty\right)\subset\mathbb{R}$
are, respectively, stiffness and viscosity parameters, $\tilde{\myvec t}_{i}^{\text{OS}}$
is the translation error of the robots in the PS, but seen from the
point of view of the OS, and $\dot{\myvec t}_{i,\text{OS}}$ is the
linear velocity of that haptic device. This proportional force feedback
with viscosity allows the operator to ``feel'' any task-space directions
in which the robot has difficulty moving. The motion-scaling (MS)
is the ratio between the motion in the PS induced by the motion in
the OS. It is important to note that in the current setup the SmartArm
is not equipped with force sensors in the instruments and cannot sense
forces applied to the environment. Further work is needed analyze
the stability of the combination of Problem~\ref{eq:quadratic_problem_teleoperation}
in the PS with the force-feedback term \eqref{eq:cartesian_impedance}
in the OS in addition to operator effects.

\subsection{Software implementation}

The software architecture is described in detail in \cite{Marinho2020}.
The computational functions to calculate the necessary Jacobians are
part of the DQ Robotics library \cite{Adorno2020}. The QP problem
was solved using IBM CPLEX\footnote{https://www.ibm.com/analytics/cplex-optimizer}.

\section{Experiments}

\textcolor{black}{}
\begin{figure}[t]
\begin{centering}
\textcolor{black}{\includegraphics[width=1\columnwidth]{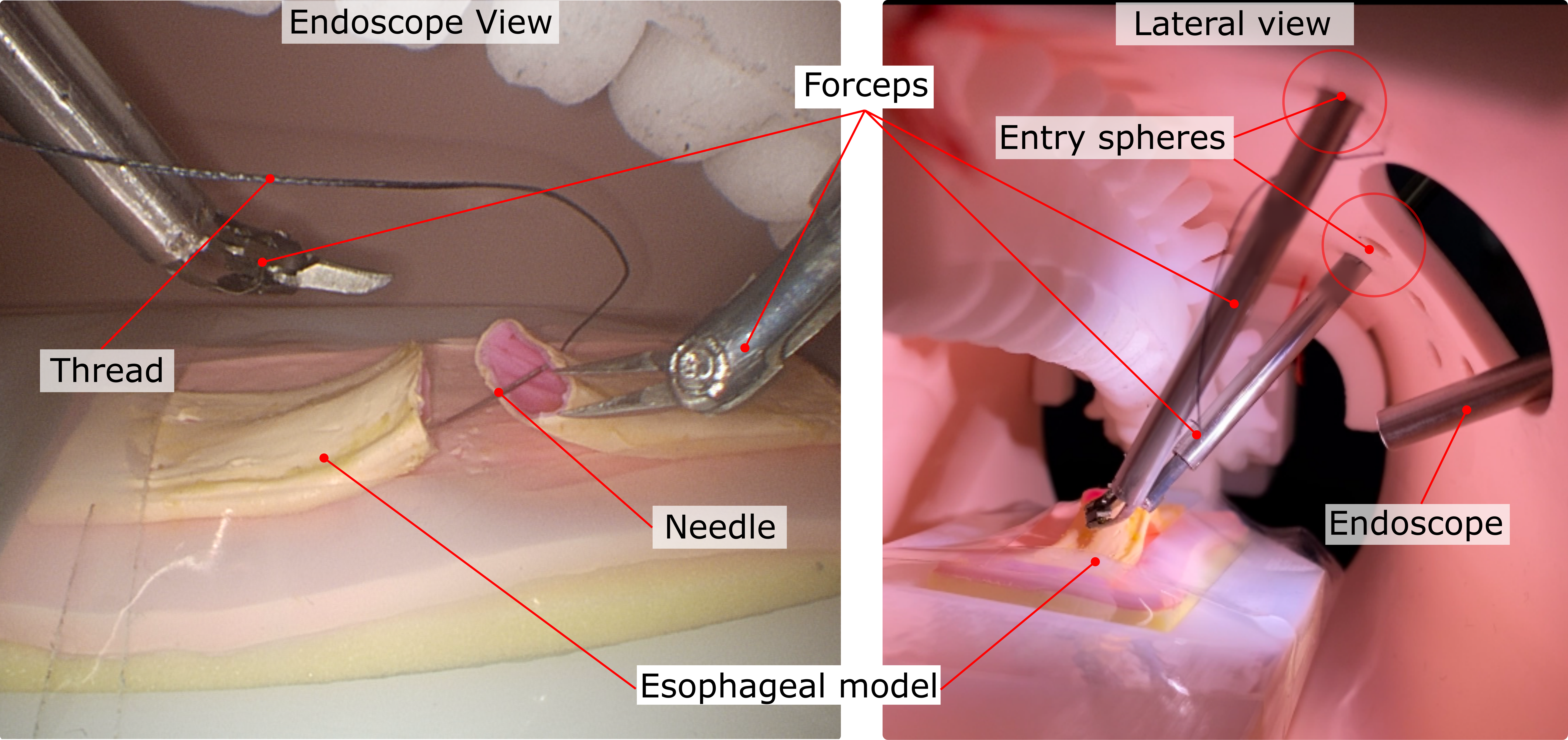}}
\par\end{centering}
\textcolor{black}{\caption{\label{fig:model_inside}Endoscopic view during operation (\emph{left}).
Lateral view showing the placement of the 3.5mm-diameter instruments,
the 4-mm-diameter endoscope, the esophageal model, and the entry spheres
(\emph{right}).}
}
\end{figure}

We designed an experiment to validate the SmartArm system in a suturing
task within the limited workspace of a neonate chest model. The chest
and esophageal models are being developed at Nagoya University by
a group of medical doctors. It is important to note that this is a
feasibility study; hence it was not designed to evaluate if the SmartArm
provides a better suturing experience than manual surgery. An external
view of the experimental setup is as shown in Fig.~\ref{fig:setup},
and a lateral view from inside the neonate model is shown in Fig.~\ref{fig:model_inside}.

An operator inexperienced in actual surgical procedures was asked
to tie, using the SmartArm system, a two-throw surgical knot connecting
both sides of the esophageal model for a total of ten trials. In each
trial, the operator handled a suture commonly used in pediatric surgery
(5-0 PERMA-HAND SILK 13 mm 3/8c, Ethicon, USA). The thread was cut
to have a length of $75$~mm.

For the purpose of time-keeping, the task started from when the operator
touched the esophageal model with the tip of the needle while holding
the needle with the robotic forceps. The task was considered to be
over when the operator finished a solid two-throw knot, that is, a
knot that could no longer be tightened in spite of pulling the thread.
This time-keeping methodology is similar to the one we used for endonasal
experiments \cite{Marinho2020} and helps to standardize the results.

\begin{table}[h]
\caption{\label{tab:PS_parameters} Parameters used on the PS and the OS.}

\centering{}%
\begin{tabular}{ccccccccc}
\hline 
\noalign{\vskip\doublerulesep}
$\alpha$ & $\eta$ & $\eta_{d}$ & $D_{\text{safe}}$ & $\lambda_{R}$ & $\lambda_{F}$ & $\eta_{f}$ & $\eta_{V}$ & MS\tabularnewline[\doublerulesep]
\hline 
\noalign{\vskip\doublerulesep}
\noalign{\vskip\doublerulesep}
0.9999 & 120 & 1 & 0.005$^{2}$ & 0.01 & 0.0 & 100 & 10 & 1\tabularnewline[\doublerulesep]
\noalign{\vskip\doublerulesep}
\end{tabular}
\end{table}

A summary of the control parameters used in this work is shown in
Table~\ref{tab:PS_parameters}. Note that the parameters were tuned
in pilot trials until the user was comfortable with the robot behavior
and kept constant through all the experimental trials reported herein.
An $\alpha\geq0.99$ is recommended for teleoperation where translation
signals are in millimeter scale compared to rotational changes that
can be in the range of several degrees. The proportional gain was
chosen as $\eta=120$ and should be set as large as possible without
causing the system to vibrate\footnote{Vibrations owing to the choice of parameters are inherent to the discrete-time
implementation and not caused by the algorithm itself.}. The entry-sphere gain was set as $\eta_{d}=1$ and can be as large
as possible without causing the system to vibrate near the constraint
boundaries. The safe distance was set to generate a 5~mm radius sphere
as indicated by medical doctors. We chose $\lambda_{F}=0.0$ in contrast
with $\lambda_{R}=0.01$ as damping factors, because in this context
it is reasonable to have the robot preferably move the forceps DoF
instead of the manipulator's DoF whenever possible. This is motivated
by the fact that the forceps DoF carry negligible weight and have
less associated inertia when compared to the manipulators' DoF.

\section{Results and discussion}

\begin{table}[tbh]
\caption{\label{tab:total_time}The total time for each suturing trial using
the SmartArm.}

\noindent\resizebox{1.0\columnwidth}{!}{%
\begin{centering}
\begin{tabular}{|c|c|c|c|c|c|c|c|c|c|c|}
\hline 
 & \textbf{T1} & \textbf{T2} & \textbf{T3} & \textbf{T4} & \textbf{T5} & \textbf{T6} & \textbf{T7} & \textbf{T8} & \textbf{T9} & \textbf{T10}\tabularnewline
\hline 
Time {[}s{]} & 1421 & 416 & 338 & 559 & 468 & 381 & \textbf{229} & \textbf{216} & 305 & 684\tabularnewline
\hline 
\end{tabular}
\par\end{centering}
}\medskip{}

In a neonate surgical scenario, an expert surgeon is expected to be
able to tie a knot in less than 5~min \cite{Harada2016}. Our medically
inexperienced operator was able to reach that mark twice, and we expect
that a medically experienced operator with proper training would have
better results.
\end{table}

\begin{figure}[t]
\begin{centering}
\includegraphics[width=1\columnwidth]{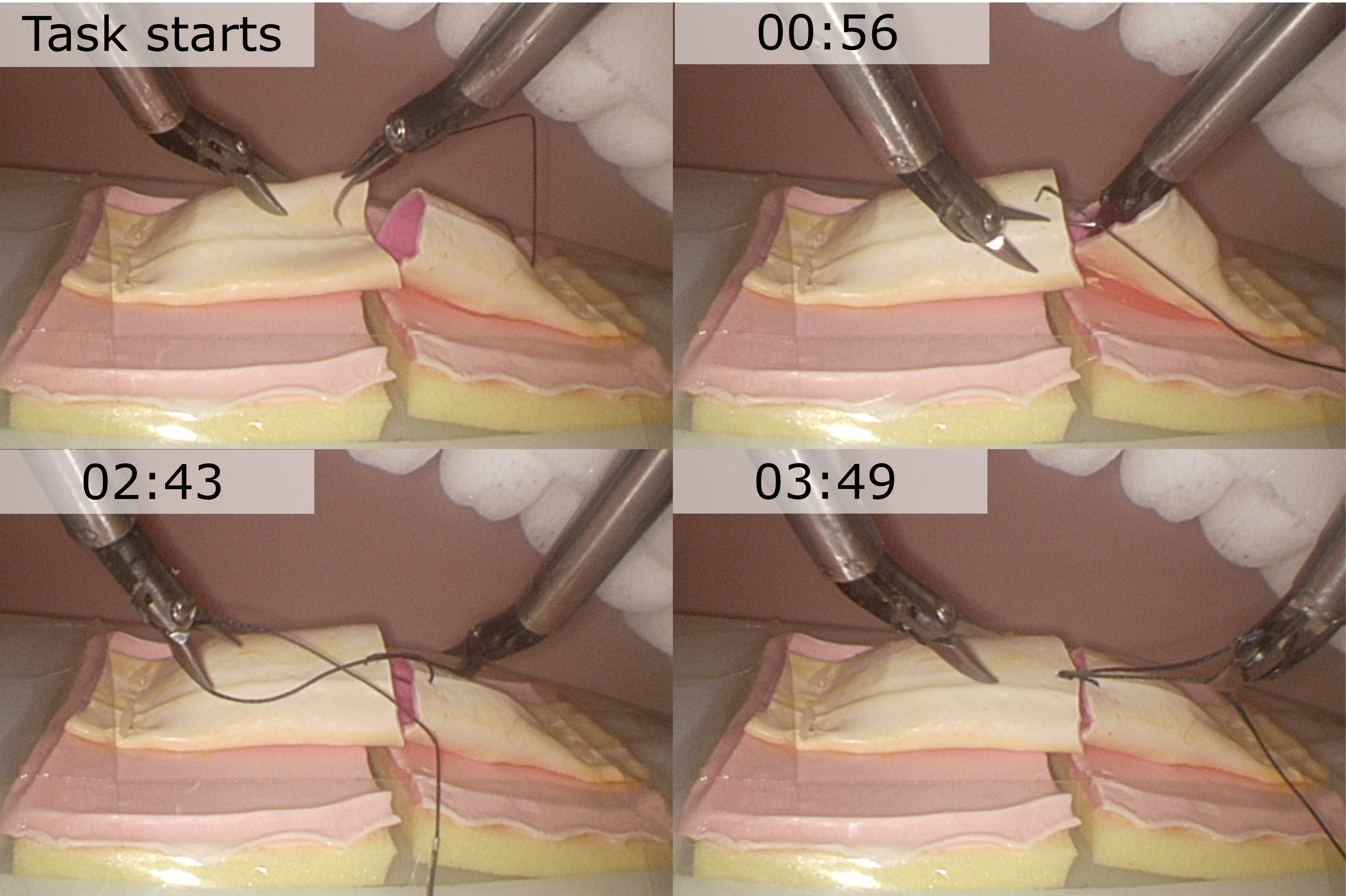}
\par\end{centering}
\caption{\label{fig:snapshots_T7}Snapshots of relevant timestamps in T7. The
first snapshot shows the needle insertion in one end of the esophageal
model. The second shows the needle insertion through the other end
of the esophageal model. The third snapshot shows the end of the first
throw. The last snapshot shows the end of the second throw.}
\end{figure}

\begin{figure}[t]
\begin{centering}
\includegraphics[width=1\columnwidth]{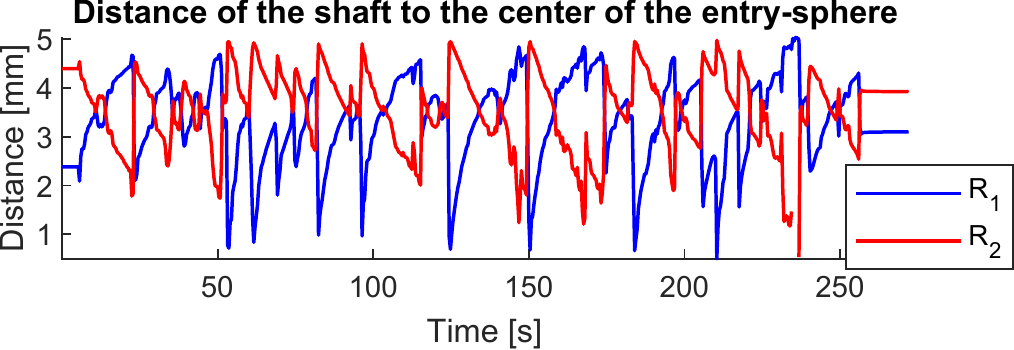}
\par\end{centering}
\caption{\label{fig:ES}Distance of the center-line of the shaft of each instruments
with respect to the center of its entry-sphere during T7 measured
using encoder information and the robot kinematic model. The distance
calculation does not take into account possible elasticity effects
of the long shaft. Note that the distance is kept within the limits
defined in Constraint~\ref{eq:constraints}.}
\end{figure}

The total time spent in each trial is summarized in Table~\ref{tab:total_time}.
Snapshots of the important timestamps of T7 are shown in Fig.~\ref{fig:snapshots_T7}.
The distance of the center-line of the shaft of each instrument with
respect to the center of its entry-sphere during T7 is shown in Fig.~\ref{fig:ES}.
The distance was measured using encoder information and the robot
kinematic model and does not take into account possible elasticity
effects of the long shaft.

The results shown in \cite[ Table 4]{Harada2016}, $T_{M}=\left\{ 189~\text{s},261~\text{s},242~\text{s},217~\text{s}\right\} $,
regard manual procedures performed by expert and qualified surgeons
in similar conditions but with different neonate chest and esophageal
models. Compared to Table~\ref{tab:total_time}, this is evidence
that suturing with the SmartArm robot inside the neonate model is
feasible within a reasonable time-frame. Prior literature \cite{Harada2016,Takazawa2018,Marinho2020}
indicates that 5~min can be considered an upper-bound for expert-level
suturing. Given the promising results in this feasibility study, we
plan on carrying this research onto the next stage and compare manual
and robot-assisted procedures in the same conditions with the participation
of surgeons.

\section{Conclusions}

In this work, we investigated the use of a versatile robotic system
for surgery, called SmartArm, in suturing experiments using neonate
esophageal and chest models. The results of this feasibility study
show, for the first time in the literature, successful robot-aided
suturing trials inside a neonate model.

Total task-time is not the only important dimension in evaluating
performance in suturing. In future work, we will evaluate the performance
of the SmartArm system when compared to the manual procedure also
in terms other important performance metrics, such as precision and
workload.

{\small{}\bibliographystyle{IEEEtran}
\bibliography{root}
}{\small\par}
\end{document}

%% file: macros.tex
\global\long\def\quat#1{\boldsymbol{#1}}%

\global\long\def\dq#1{\underline{\boldsymbol{#1}}}%

\global\long\def\hp{\mathbb{H}_{p}}%

\global\long\def\dotmul#1#2{\langle#1,#2\rangle}%

\global\long\def\partialfrac#1#2{\frac{\partial\left(#1\right)}{\partial#2}}%

\global\long\def\totalderivative#1#2{\frac{d}{d#2}\left(#1\right)}%

\global\long\def\mymatrix#1{\boldsymbol{#1}}%

\global\long\def\vecthree#1{\operatorname{vec}_{3}#1}%

\global\long\def\vecfour#1{\operatorname{vec}_{4}#1}%

\global\long\def\haminuseight#1{\overset{-}{\mymatrix H}_{8}\left(#1\right)}%

\global\long\def\hapluseight#1{\overset{+}{\mymatrix H}_{8}\left(#1\right)}%

\global\long\def\haminus#1{\overset{-}{\mymatrix H}_{4}\left(#1\right)}%

\global\long\def\haplus#1{\overset{+}{\mymatrix H}_{4}\left(#1\right)}%

\global\long\def\norm#1{\left\Vert #1\right\Vert }%

\global\long\def\abs#1{\left|#1\right|}%

\global\long\def\conj#1{#1^{*}}%

\global\long\def\veceight#1{\operatorname{vec}_{8}#1}%

\global\long\def\myvec#1{\boldsymbol{#1}}%

\global\long\def\imi{\hat{\imath}}%

\global\long\def\imj{\hat{\jmath}}%

\global\long\def\imk{\hat{k}}%

\global\long\def\dual{\varepsilon}%

\global\long\def\getp#1{\operatorname{\mathcal{P}}\left(#1\right)}%

\global\long\def\getpdot#1{\operatorname{\dot{\mathcal{P}}}\left(#1\right)}%

\global\long\def\getd#1{\operatorname{\mathcal{D}}\left(#1\right)}%

\global\long\def\getddot#1{\operatorname{\dot{\mathcal{D}}}\left(#1\right)}%

\global\long\def\real#1{\operatorname{\mathrm{Re}}\left(#1\right)}%

\global\long\def\imag#1{\operatorname{\mathrm{Im}}\left(#1\right)}%

\global\long\def\spin{\text{Spin}(3)}%

\global\long\def\spinr{\text{Spin}(3){\ltimes}\mathbb{R}^{3}}%

\global\long\def\distance#1#2#3{d_{#1,\mathrm{#2}}^{#3}}%

\global\long\def\distancejacobian#1#2#3{\boldsymbol{J}_{#1,#2}^{#3}}%

\global\long\def\distancegain#1#2#3{\eta_{#1,#2}^{#3}}%

\global\long\def\distanceerror#1#2#3{\tilde{d}_{#1,#2}^{#3}}%

\global\long\def\dotdistance#1#2#3{\dot{d}_{#1,#2}^{#3}}%

\global\long\def\distanceorigin#1{d_{#1}}%

\global\long\def\dotdistanceorigin#1{\dot{d}_{#1}}%

\global\long\def\squaredistance#1#2#3{D_{#1,#2}^{#3}}%

\global\long\def\dotsquaredistance#1#2#3{\dot{D}_{#1,#2}^{#3}}%

\global\long\def\squaredistanceerror#1#2#3{\tilde{D}_{#1,#2}^{#3}}%

\global\long\def\squaredistanceorigin#1{D_{#1}}%

\global\long\def\dotsquaredistanceorigin#1{\dot{D}_{#1}}%

\global\long\def\crossmatrix#1{\overline{\mymatrix S}\left(#1\right)}%

\global\long\def\constraint#1#2#3{\mathcal{C}_{\mathrm{#1},\mathrm{#2}}^{\mathrm{#3}}}%